# Word Embeddings and Validity Indexes in Fuzzy Clustering

Danial Toufani-Movaghar
*Computer Intelligent System Laboratory,*

*Department of Computer Engineering*
*University of Tabriz*
*Tabriz, Iran*

d.toofani@tabrizu.ac.ir

Mohammad-Reza Feizi-Derakhshi
*Computer Intelligent System Laboratory,*
*Department of Computer Engineering*
*University of Tabriz*
*Tabriz, Iran*

mfeizi@tabrizu.ac.ir

*Abstract* – In the new era of internet systems and applications, a concept of detecting distinguished topics from huge amounts of text has gained a lot of attention. These methods use representation of text in a numerical format – called embeddings – to imitate human-based semantic similarity between words. In this study, we perform a fuzzy-based analysis of various vector representations of words, i.e., word embeddings. Also we introduce new methods of fuzzy clustering based on hybrid implementation of fuzzy clustering methods with an evolutionary algorithm named Forest Optimization. We use two popular fuzzy clustering algorithms on count-based word embeddings, with different methods and dimensionality. Words about covid from Kaggle dataset gathered and calculated into vectors and clustered. The results indicate that fuzzy clustering algorithms are very sensitive to high-dimensional data, and parameter tuning can dramatically change their performance. We evaluate results of experiments with various clustering validity indexes to compare different algorithm variation with different embeddings accuracy.

*Keywords* -- fuzzy clustering, fuzzy C-means, fuzzy Gustafson-Kessel, Forest Optimization, cluster validity index, word embeddings, word vectors

I. INTRODUCTION

Word embeddings become an essential element of methods that focus on analysis and comparison of texts. Some of the most popular embeddings are *Word2Vec*, *GloVe, T5 and Bert*. The embedding is obtained via analysis of word-word co-occurrences in a text corpus. A natural question is related to the ability of the embeddings to represent a human-based semantic similarity of words. We showed that embedding has a significant effect on the result of clustering [1].

Data clustering is the process of grouping objects in a way that similarity between data points that belong to the same group (cluster) becomes as high as possible, while similarity between points from different groups gets as small as possible. It is an important task in analysis processes and has been successfully applied to pattern recognition, image segmentation, and fault diagnosis and search engines [1]. Fuzzy clustering which allows data points to belong to several numbers of clusters with different membership grades has been proved to have useful applications in many areas. Specifically, fuzzy C-Means (FCM) clustering and its augmented version – fuzzy Gustafson-Kessel (FGK) clustering are the most popular fuzzy clustering techniques. Both of the methods suffer from Termination in local optima and mixing these useful methods with an evolutionary meta-heuristic search method like Forest Optimization Algorithm guarantees avoiding such a situation. High-dimensional spaces often have a devastating effect on data clustering in terms of performance and quality; this issue is regarded as the *curse of dimensionality*. Fuzzy Gustafson-Kessel algorithm suffers from performance in large dimensional data.

Our study evaluates 4 method of clustering: Fuzzy C-Mean, Fuzzy Gustafson-Kessel, FOA Fuzzy C-Mean witch is hybrid implementation of Fuzzy C-Mean and Forest Optimization, and FOA Gustafson-Kessel which is hybrid implementation of Fuzzy Gustafson Kessel and Forest Optimization. We evaluated results on a text dataset with 500 items that were embedded with 4 different algorithms and they have the same output dimension.
Additionally, we illustrate 'usefulness' of fuzzy clustering via analysis of degrees of belongings of words to different clusters. For comparison of embedding and various algorithms involved we used 7 clustering validity indexes which were described in further sections.

The paper is structured as follows: Section II contains the description of Fuzzy C-Means (FCM) and Fuzzy Gustafson-Kessel (FGK) algorithms and their evolutionary hybrid version with Forest Optimization (FOA). In Section III, we provide an overview of the validity indices applied for fuzzy clustering processes. Then in section IV we describe briefly embedding methods used in this paper. Section V shows our experimental results. Finally, the obtained conclusions are presented in Section VI.

II. FUZZY CLUSTERING AND THEIR HYBRID VERSION

In contrast to hard clustering techniques, where one point is assigned exactly to only one cluster, fuzzy clustering allows data points to pertain to several clusters with different grades of membership [1].

We have analyzed the behavior of FCM and FGK in clustering of vector representations of words in different dimensions. The details of these clustering methods have been described in the following sections.

*A. Fuzzy C-means clustering*

The Fuzzy C-means algorithm [1], allows an observation to belong to multiple clusters with varying grades of membership. Having D as the number of data points, $N$ as the number of clusters, $m$ as the fuzzifier parameter, as the $i$-th data point, as the center of the $j$-th cluster, as the membership degree of for the $j$-th cluster, FCM aims to minimize :

$$J = \sum_{i=1}^{D} \sum_{j=1}^{N} \mu_{ij}^m \left| x_i - c_j \right|^2 \quad (1)$$

The FCM clustering proceeds in following way [1] :
1. Cluster membership values $\mu_{ij}$ and initial cluster centers are initialized randomly.
2. Cluster centers are computed according to the formula:

$$c_j = \frac{\sum_{i=1}^{D} \mu_{ij} x_i}{\sum_{i=1}^{D} \mu_{ij}^m}$$

3. Membership graded $\mu_{ij}$ are updated and The objective function $J$ is calculated
4. The steps 2, 3, 4 are repeated until the value of the objective function gets less than a specified threshold.

*B. Fuzzy Gustafson-Kessel clustering*

Fuzzy Gustafson-Kessel (FGK) extends FCM by introducing an adaptive distance norm that allows the algorithm to identify clusters with different geometrical shapes [1]. The distance metric is defined as it should be computed from a fuzzy covariance matrix of each cluster:

$$D_{GK}^2 = (x_k - V_i)^T A_i (x_k - V_i)$$

Where $A_i$ is computed from a fuzzy covariance matrix of each cluster:

$$A_i = (\rho_i |C_i|)^{\frac{1}{d}} C_i^{-1}$$

$$c_i = \frac{\sum_{k=1}^{N} \mu_{ik}^m (x_k - V_i)^T (x_k - V_i)}{\sum_{k=1}^{N} \mu_{ik}^m}$$

Enabling the matrix to change with fixed determinant serves to optimize the shape of clusters by keeping the cluster's volume constant [1]. Gustafson-Kessel clustering minimizes the following criterion:

$$J = \sum_{i=1}^{c} \sum_{k=1}^{N} \mu_{ik}^m D_{GK}^2 = \sum_{i=1}^{c} \sum_{i=1}^{N} \mu_{ik}^m (x_k - v_i)^T A_i (x_k - v_i)$$

*C. Forest Optimization Algorithm*

The Forest optimization algorithm (FOA) (Ghaemi and Feizi-Derakhshi, 2014) is suitable for continuous non-linear optimization problems. The algorithm is inspired by the existence of ancient trees after decades. While many of the trees are short-lived, the number of trees is still in existence even after a few decades. In this algorithm, spreading seeds of the trees are simulated so that the number of seeds is set under the trees, and the number of seeds spread in a vast area of Forest by natural events such as wind. The output of the algorithm suggests improving its accuracy in finding the optimal positions rather than genetic algorithm and particle swarm optimization [2].

This algorithm has three main steps: (1) local seed production, (2) removing some members of the population, (3) global seed production. In this algorithm, like other evolutionary algorithms, forest trees are initialized in the initialization step. A tree, in addition to the values of the variables, has a part that represents the age of the related tree. The age of each tree is zero at first. After initialization of the trees, in the local seed production step, some new trees aged zero (new seeds) are created. Then, one unit is added to the age of previous trees. In the second step, the number of trees based on the pre-defined population, should be removed from the population.

Removing the excess trees is based on their fitness function values. Therefore some trees will be omitted from the forest and they will form the candidate population for the global seeding step. In the production of global seeds, a percentage of the candidate population is chosen to move far in the forest. Global seeding step adds some new potential solutions to the forest so that it goes away from local optimums. Finally, after sorting the trees according to their fitness value, the tree with the highest fitness value is selected as the best tree. Then the age of the best tree will be set to 0 in order to avoid the aging of the best tree as the result of a local seeding step.

*D. Hybrid Algorithm*

In this section we introduce our work as a new method for clustering as a combination of Fuzzy C Mean and Fuzzy Gustafson Kessel with Forest Optimization Algorithm. In both fuzzy c mean and Gustafson Kessel we initialize some random vector as the center of clusters and then in a local search process we minimize the objective function to find the best center for clusters.

Instead in our proposed hybrid algorithm lets forest optimization manage local and global search and initiate cluster centers and put cluster center calculation to original algorithm with single iteration, also because both algorithm objective functions are target to minimization of some function, we inject this as fitness function of Forrest optimization. Therefore, forest optimization does both local and global search and calculations were done by fuzzy c mean and Gustafson Kessel.

## III Validity Index

There are several validity indices to analyze the performance of the fuzzy clustering algorithms. Each of them targets one or more aspect of clustering quality. Table below shows 7 validity indexed we used in our study and a brief description about them [3, 4].

| Title | Formula | Description |
|---|---|---|
| Partition Coefficient (PC) | $PC = \frac{1}{N} \sum_{k=1}^{N} \sum_{i=1}^{c} \mu_{ik}^{2}$ | PC changes between [0,1] range and the maximum value indicates the best clustering quality |
| Normalized Partition Coefficient (NPC) | $V_{NPC}(U,c) = 1 - \frac{c}{c-1}(1 - V_{PC}(U,c))$ $= 1 - \frac{c}{c-1}\left(1 - \frac{1}{n}\sum_{k=1}^{n}\sum_{i=1}^{c}\mu_{ik}^{2}\right)$ | Normalized version of PC |
| FHV (Fuzzy Hyper Volume) | $FHV = \sum_{k=1}^{C}[\det(F_k)]^{\frac{1}{2}}, \; F_k = \frac{\sum_{i=1}^{N}\mu_{ki}^{m}(x_i - v_k)(x_i - v_k)^T}{\sum_{i=1}^{N}\mu_{ki}^{m}}$ | Recognize clusters of different sizes independent of their expansion, their location in the feature space, and their closeness to each other |
| FS (Fukuyama And Sageo) | $FS = J_m(u,v) - K_m(u,v)$ $= \sum_{k=1}^{C}\sum_{i=1}^{N}\mu_{ki}^{m} \| x_i - v_k \|^{2} - \sum_{k=1}^{C}\sum_{i=1}^{N}\mu_{ki}^{m} \| v_k - v^{-} \|^{2}$ | Measures the distances between data items and cluster prototypes directly, the separation measure measures the distances between cluster centers and the grand mean of the data set |
| XB (Xie and Beni) | $XB = \sum_{i=1}^{c} \frac{\sum_{k=1}^{N}(\mu_{ik})^m \| x_k - v_i \|^{2}}{N\min_{ik} \| x_k - V_i \|^{2}}$ | Favors partitionings where all clusters are well separated from each other |
| Beringer-Hullermeier Index (BH) | $V_{BH}(U,X,V) = \frac{1}{n}\sum_{i=1}^{c}\sum_{k=1}^{n}\mu_{ik}^{m}\|x_k - v_i\|^{2} \times \sum_{i=1}^{c-1}\sum_{j=i+1}^{c}\frac{1}{D(C_i,C_j)}$ | Measures the compactness within clusters as intra-cluster similarity and the overlap |

| Bouguessa-Wang-Sun index (BWS) | $Comp_{BWS}(U,V,X) = \sum_{i=1}^{c} tr(Cov_i)$ $Sep_{BWS}(U,V,X) = tr\left(\sum_{i=1}^{c}\sum_{k=1}^{n} \mu_{ik}^m (v_i - \bar{x})(v_i - \bar{x})^T\right)$ with $\bar{x} = \frac{\sum_{k=1}^{n} x_k}{n}$ $V_{BWS}(U,V,X) = \frac{Sep_{BWS}(U,V,X)}{Comp_{BWS}(U,V,X)}$ | between clusters as inter-cluster similarity |
|---|---|---|
| | | Analyzes partitionings of the data set particularly with regard to the overlaps between clusters and the variations in the cluster density, orientation and shape |

Table 1: Clustering validity indexes formulas

IV Embedding's

A word embedding is a learned representation for text where words that have the same meaning have a similar representation. Various embedding methods exist and each represent a topic as a vector of numbers with different dimensions. In This study we used 4 known Embedding methods named Word2Vec, Bert, Gloves and T5.

*A. Word2Vec Embedding*

Word2vec represents words in vector space representation. Words are represented in the form of vectors and placement is done in such a way that similar meaning words appear together and dissimilar words are located far away. This is also termed as a semantic relationship. Neural networks do not understand text, instead they understand only numbers. Word Embedding provides a way to convert text to a numeric vector.

Word2vec reconstructs the linguistic context of words. Before going further let us understand, what is the linguistic context? In general life scenarios when we speak or write to communicate, other people try to figure out what is the objective of the sentence. For example, "What is the temperature of India", here the context is the user wants to know "temperature of India" which is context. In short, the main objective of a sentence is context. Word or sentence surrounding spoken or written language (disclosure) helps in determining the meaning of context. Word2vec learns vector representation of words through the contexts [1].

*B. BERT Embedding*

BERT, which stands for Bidirectional Encoder Representations from Transformers. Unlike recent language representation models, BERT is designed to pre-train deep bidirectional representations from unlabeled text by jointly conditioning on both left and right context in all layers. As a result, the pre-trained BERT model can be fine-tuned with just one additional output layer to create state-of-the-art models for a wide range of tasks, such as question answering and language inference, without substantial task-specific architecture modifications.

*C. Glove Embedding*

GloVe is essentially a log-bilinear model with a weighted least-squares objective. The main intuition underlying the model is the simple observation that ratios of word-word co-occurrence probabilities have the potential for encoding some form of meaning. For example, consider the co-occurrence probabilities for target words ice and steam with various probe words from the vocabulary. Here are some actual probabilities from a 6 billion word corpus:

| Probability and Ratio | $k$ = solid | $k$ = gas | $k$ = water | $k$ = fashion |
|---|---|---|---|---|
| $P(k \mid \text{ice})$ | ١.٩ × ١٠⁻⁴ | ٦.٦ × ١٠⁻⁵ | ٣.٠ × ١٠⁻³ | ١.٧ × ١٠⁻⁵ |
| $P(k \mid \text{steam})$ | ٢.٢ × ١٠⁻⁵ | ٧.٨ × ١٠⁻⁴ | ٢.٢ × ١٠⁻³ | ١.٨ × ١٠⁻⁵ |
| $P(k \mid \text{ice})/P(k \mid \text{steam})$ | ٨.٩ | ٨.٥ × ١٠⁻² | ١.٣٦ | ٠.٩٦ |

As one might expect, ice co-occurs more frequently with solid than it does with gas, whereas steam co-occurs more frequently with gas than it does with solid. Both words co-occur with their shared property water frequently, and both co-occur with the unrelated word fashion infrequently. Only in the ratio of probabilities does noise from non-discriminative words like water and fashion cancel out, so that large values (much greater than 1) correlate well with properties specific to ice, and small values (much less than 1) correlate well with properties specific to steam. In this way, the ratio of probabilities encodes some crude form of meaning associated with the abstract concept of thermodynamic phase.

The training objective of GloVe is to learn word vectors such that their dot product equals the logarithm of the words' probability of co-occurrence. Owing to the fact that the logarithm of a ratio equals the difference of logarithms, this objective associates (the logarithm of) ratios of co-occurrence probabilities with vector differences in the word vector space. Because these ratios can encode some form of meaning, this information is encoded as vector differences as well. For this reason, the resulting word vectors perform very well on word analogy tasks, such as those examined in the word2vec package.

*D. T5 Embedding*

T5 is a recently released encoder-decoder model that reaches SOTA results by solving NLP problems with a text-to-text approach. This is where text is used as both an input and an output for solving all types of tasks. This was introduced in the recent paper, Exploring the Limits of Transfer Learning with a Unified Text-to-Text Transformer (paper). Some unique features of T5 embedding are as follows:
- Encoder-decoder Transformer
- Relative Positional Self-Attention
- $S\,rel$ is the simplified positional embedding

The offset between key and query being compared in the self-attention mechanism. Each word embedding is a scalar with a position parameter which is shared across model layers. Some pros and cons of model listed below:
- "text-to-text" format
- consistent training objective: maximum likelihood
- task-specific (text) prefix
- Mismatch label Issue

V Experiments

For evaluation of a defined algorithm, we choose 2 datasets. For testing and proof of concept we used IRIS dataset which led to the results in Figure 1. The results of 4 fuzzy algorithms explained in section II are shown in table. For validity of clustering we used 7 different clustering validity index listed in section III. In execution of algorithm we run these algorithm for cluster count 1 to 5 and gather validity index values for each of clusters. Finally according to Validity Index formula if it needs to be Minimized or Maximized, We choose the best value for it.

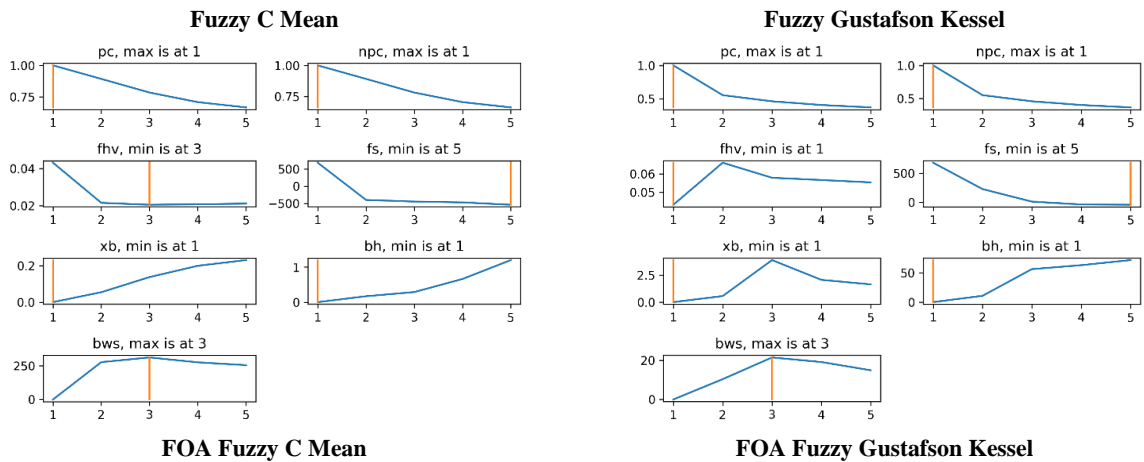

**FOA Fuzzy C Mean**  **FOA Fuzzy Gustafson Kessel**

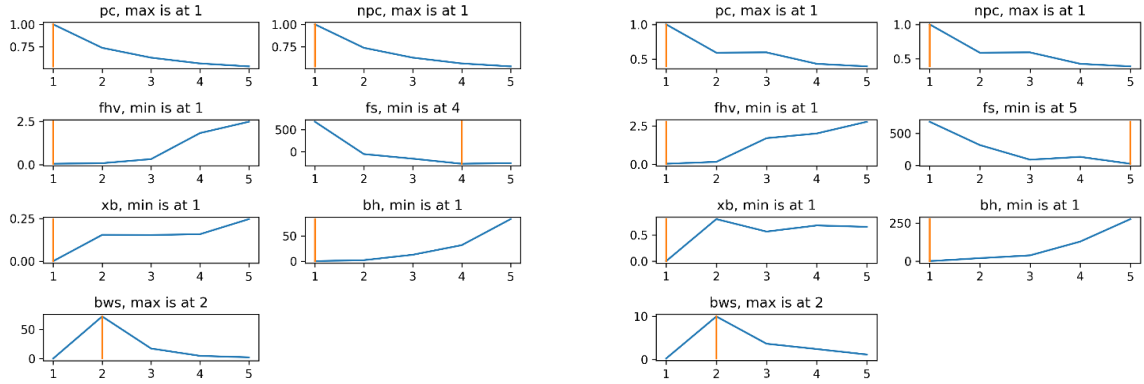

Figure 1: Iris dataset results

In comparison of running time of algorithms, we run FOA version with 10 iteration count (epochs), so each iteration spent time equivalent to a run of its base version, therefore running time in FOA version of algorithms cost N time slower than basic version, as N is the number of iteration we selected for FOA.

| Algorithm | Run Time (Seconds) |
|---|---|
| *Fuzzy C Mean* | 20 |
| *Fuzzy Gustafson Kessel* | 63 |
| *FOA Fuzzy C Mean* | 315 |
| *FOA Fuzzy Gustafson Kessel* | 970 |

Table 2: Running Time Comparison of Algorithms

As we know IRIS is a 3 cluster dataset, so functionalities of algorithms are identified if it detects 3 as result of one or more of CVI calculated. In our result we evaluated algorithms for cluster count 1 to 5. If result is Near Correct number of cluster with variance of 1 (in IRIS 2 or 4) then we count it in second column.

| Algorithm | No of correct detection | No of Near Correct detection |
|---|---|---|
| *Fuzzy C Mean* | 2 | 0 |
| *Fuzzy Gustafson Kessel* | 1 | 0 |
| *FOA Fuzzy C Mean* | - | 2 |
| *FOA Fuzzy Gustafson Kessel* | - | 1 |

Table 3: Correct detected no of clusters for algorithms

As result shows, hybrid algorithm in addition to suffering from poor performance and long running time, because of searching iteratively in dataset for finding the best center, detects wrong number of clusters and approximately detects correct not of clusters. So in the rest of research we ignore hybrid version of algorithm and gather results for basic version of algorithms.

For real comparison of metrics provided in this article, we used two large text dataset with is gathered from Twitter about COVID-19 in interval from 29 Mars 2020 through 30 April 2020 [6, 7]. This data set contains data from users which applies following hashtags in their comments:

#coronavirus, #coronavirusoutbreak, #coronavirusPandemic, #covid19, #covid_19, #epitwitter, #ihavecorona, #StayHomeStaySafe, #TestTraceIsolate

Data contains tweet content, Account twitted, Hashtags and location. Retweet argument is set to FALSE and they will be ignored. Dataset language is set to 'EN' and twits in English language is regarded. For folding of data we only suppose data from Mondays of each week in our survey and in each Monday we choose 100 twits. Therefore, in these 5 week we have 500 record for analysis. In text pre-processing all Emoji's, Mentions, Links and Hashtags are removed and after tokenization if twit remains with just 4 words, this twit is removed and replaced with a new one. After all data is embedded with 4 algorithm named, Word2Vec, Glove, Bert and T5. Then results are fed into both Fuzzy C Mean and Fuzzy Gustafson Kessel algorithm and results are gathered in the below figure 2.

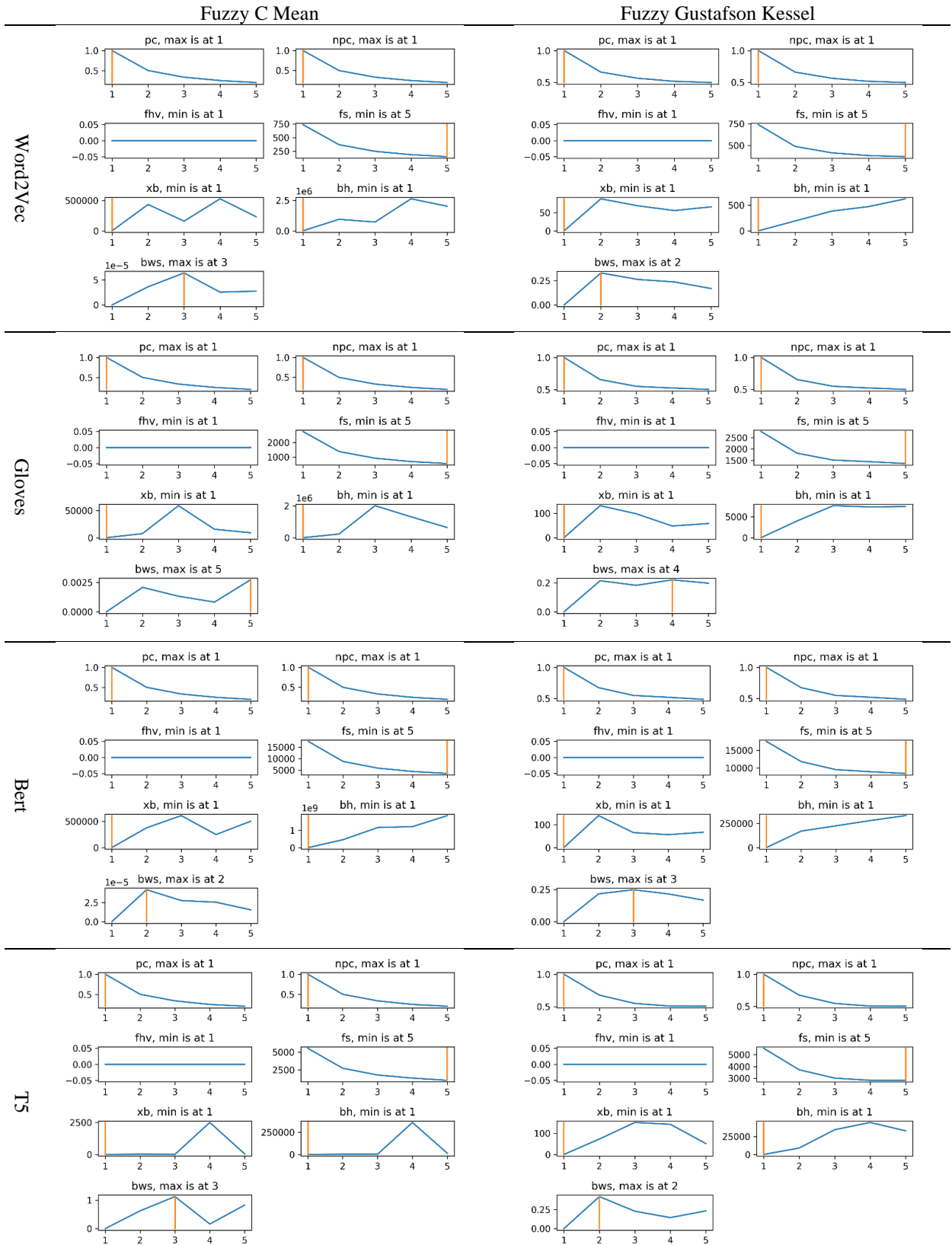

Figure 2: Topic Clustering result for Covid-19 Dataset

For comparison of Embedding's result, data gathered in this manner. If result of embedding detected correct number of clusters in one of algorithms, then we count it in No of Correct Detection. If result is Near Correct number of cluster with variance of 1 then we count it in second column.

| Embedding | No of Correct Detection | No Of Near Correct Detection |
|---|---|---|
| *Word2Vec* | 1 | 1 |
| *Gloves* | 0 | 1 |
| *Bert* | 1 | 1 |
| *T5* | 1 | 1 |

Table 4: Comparison of results based on Embedding

For comparison of Validity Indexes, we gather data in below table with this manner. If Index detect the correct number of clusters then in that column we count it, also if index detect one below or top of number of clusters we count that in second column.

| Validity Index | No of Correct Detection | No Of Near Correct Detection |
|---|---|---|
| *Partition Coefficient (PC)* | 0 | 0 |
| *Normalized Partition Coefficient (NPC)* | 0 | 0 |
| *Fuzzy Hypervolume (FHV* | 0 | 0 |
| *Fukuyama-Sugeno Index (FS)* | 0 | 0 |
| *Xie-Beni Index (XB)* | 0 | 0 |
| *Beringer-Hullermeier Index (BH)* | 0 | 0 |
| *Bouguessa-Wang-Sun index (BWS)* | 4 | 3 |

Table 5: Comparison of Validity Indexes

VI Results

In execution of algorithms defined in section II, because of calculation of covariance matrix, Fuzzy Gustafson Kessel Algorithm is 10 time slower than Fuzzy C Mean. Also in proposed hybrid form of algorithm because of using evolutionary algorithm which searches all around dataset, calculation time exceed and the time of execution is more than normal fuzzy algorithm. Also with high dimensional data, calculation of covariance matrix for Fuzzy Gustafson Kessel is time consuming and is not recommended. Because of high dimension of text embedding results, we do not propose Fuzzy Gustafson Kessel for such application, also in our result we eliminated calculation of results for hybrid version of Fuzzy Gustafson Kessel.

In Table 2, evaluation of cluster validity indexes in 4 different embedding described in section IV are shown. With Word2Vec, Glove and Bert embedding with all two algorithm shown, As result shows other embedding but Glove do their jobs very well and help in clustering of large text dataset. But according to results, T5 and Word2Vec embedding with execution of both algorithms detect correct number of clusters.

Among validity Indexes in text clustering just Bouguessa-Wang-Sun index (BWS) from the 7 validity index has ability to detect right number of clusters in big text data. So our proposed method in clustering is using Bouguessa-Wang-Sun index (BWS).

## VI Conclusion

In topic clustering in addition to accuracy, time to evaluate results are a main factor. Fuzzy C Mean as a basic and useful algorithm is the proposed algorithm for doing job. Also embedding is a main factor which led to more accurate result plays an important role. One of T5, Bert of Word2Vec Embeddings can be used and our proposed technique for embedding is T5. Among validity indexes Bouguessa-Wang-Sun index (BWS) is Best cluster validity index according to results.

## VII References

1. Atakishiyev S., Reformat M.Z. (2021) Analysis of Word Embeddings Using Fuzzy Clustering.
2. M Ghaemi, MR Feizi-Derakhshi - Expert Systems with Applications, 2014
3. Ryoo, J.H.; Park, S.; Kim, S.; Ryoo, H.S. Efficiency of Cluster Validity Indexes in Fuzzy Clusterwise Generalized Structured Component Analysis. Symmetry 2020, 12, 1514.
4. Himmelspach L., Conrad S. (2010) Fuzzy Clustering of Incomplete Data Based on Cluster Dispersion. In: Hüllermeier E., Kruse R., Hoffmann F. (eds) Computational Intelligence for Knowledge-Based Systems Design. IPMU 2010.
5. Liu, Yongli & Zhang, Xiaoyang & Chen, Jingli & Chao, Hao. (2019). A Validity Index for Fuzzy Clustering Based on Bipartite Modularity. Journal of Electrical and Computer Engineering. 2019. 1-9. 10.1155/2019/2719617.
6. https://www.kaggle.com/smid80/coronavirus-covid19-tweets-early-april
7. https://www.kaggle.com/smid80/coronavirus-covid19-tweets-late-april